\documentclass{article} 
\usepackage{iclr2027_conference,times}

\usepackage{amsmath,amsfonts,bm}

\def\eqref#1{equation~\ref{#1}}

\def\1{\bm{1}}

\DeclareMathAlphabet{\mathsfit}{\encodingdefault}{\sfdefault}{m}{sl}
\SetMathAlphabet{\mathsfit}{bold}{\encodingdefault}{\sfdefault}{bx}{n}

\usepackage{hyperref}
\usepackage{url}
\usepackage{graphicx}
\usepackage{subcaption}
\usepackage{algorithm}
\usepackage{algorithmic}
\usepackage{xcolor}
\usepackage{booktabs}
\usepackage{multirow}
\usepackage{multicol}
\usepackage{float}
\title{APEX: Speculate Smarter, Not Deeper}

\author{
Manvi Jha\thanks{Work was done during an internship at AWS AI.} \\
University of Illinois Urbana-Champaign \\
AWS AI \\
\texttt{manvij2@illinois.edu}
\And
Zach Zhang \\
AWS AI \\
\texttt{zhiang@amazon.com}
\And
Zhichao Xu \\
AWS AI
\And
Linbo Liu \\
AWS AI
\And
Sai Muralidhar Jayanthi \\
AWS AI
\And
Vinayak Arannil \\
AWS AI \\
\texttt{varannil@amazon.com}
}

\usepackage{xspace}
\usepackage{tikz}
\usetikzlibrary{tikzmark}
\makeatletter
\newcommand*\myfontsize{%
  \@setfontsize\myfontsize{6.7}{8}%
}
\makeatother
\newcommand{\mytextbox}[2]{\tikzmarknode[draw=#1,thick,inner sep=2pt]{test}{\myfontsize #2}}

\definecolor{cadmiumgreen}{rgb}{0.0, 0.42, 0.24}

\definecolor{myred}{rgb}{0.7, 0.3, 0.0}
\definecolor{myblue}{rgb}{0.2, 0.3, 0.6}
\newcommand{\apex}{\mytextbox{myred}{\textbf{\textcolor{myred}{APEX}}}\xspace}
\newcommand{\depth}{\mytextbox{myblue}{\textbf{\textcolor{myblue}{APEX-Depth}}}\xspace}
\newcommand{\router}{\mytextbox{myblue}{\textbf{\textcolor{myblue}{APEX-Router}}}\xspace}

\iclrfinalcopy 
\begin{document}

\maketitle


\begin{abstract}

Speculative decoding reduces large language model inference latency by drafting multiple tokens before target-model verification, but its effectiveness depends on both the proposal mechanism and draft depth. Fixed configurations cannot respond to changes in predictability, repetition, and acceptance during generation, so deeper drafting can increase wasted computation without proportional speedup. We introduce \apex, a learned controller that balances decoding speed and draft-token waste through request-level expert selection and block-level depth adaptation. \router selects among EAGLE-3, n-gram, and draft-model speculation for each request, while \depth adjusts draft length at each verification block using causal decoding signals and recent verifier feedback. \apex models accepted draft length as censored survival feedback, learning position-wise rejection hazards, block execution costs, and an action utility that balances throughput, accepted progress, and wasted tokens. This allows the controller to adapt speculation while retaining the target model’s verification procedure. We integrate \apex into vLLM and evaluate it with Qwen3-8B across six workloads, achieving up to $5.24\times$ speedup over autoregressive decoding. Across the aggregate evaluation, \apex-S achieves $4.27\times$ speedup, while \apex-B achieves $3.27\times$ speedup with a $41.0\%$ relative reduction in wasted-token percentage compared with fixed n-gram speculation at $k=16$, providing distinct operating points for balancing acceleration and draft-token utilization.
\end{abstract}

\section{Introduction}
Large language models (LLMs) increasingly serve interactive and long-form workloads such as code generation, mathematical reasoning, and conversational assistance. In these settings, inference latency is not merely a backend cost, it directly affects user-perceived responsiveness and the feasibility of deploying capable models at scale~\cite{kwon2023vllm,agrawal2024sarathiserve}. Yet acceleration must preserve the behavior of the target model. The systems challenge is therefore not simply to generate more tokens per second, but to convert additional computation into verified output progress without changing the target model's generation semantics.

Autoregressive decoding~\cite{schuurmans2024autoregressivelargelanguagemodels} produces each token conditioned on the preceding prefix. Speculative decoding (SD)~\cite{xia-etal-2023-speculative} reduces the number of sequential target-model invocations by using a cheaper mechanism to propose multiple tokens, which the target model then verifies. Accepting a long draft prefix allows generation to advance several positions in one verification step. However, an early rejection leaves much of the proposed draft unused. Increasing draft depth consequently offers more potential progress while also increasing the amount of speculative work exposed to rejection. High speedup can coexist with substantial draft-token waste, and low waste can coexist with slow execution. Both wall-clock latency and draft-token utilization are therefore needed to assess the resulting trade-off.

The proposal mechanism also shapes this trade-off. Learned speculators such as EAGLE-3~\cite{li2025eagle3scalinginferenceacceleration} can provide useful drafts when upcoming tokens are predictable. Retrieval-based approaches like n-gram reuse available text~\cite{he2024restretrievalbasedspeculativedecoding}, making inexpensive matching methods attractive when repeated structure is present. A separate draft model can produce plausible continuations across varied inputs, but its execution cost may offset the benefit of accepting additional tokens. These mechanisms therefore differ in both the drafts they produce and the cost of producing them. The same acceptance rate or wasted-token percentage can have different latency consequences across mechanisms.

The value of speculation depends jointly on the proposal mechanism, draft depth, and decoding state. A mechanism that proposes tokens cheaply may still discard a large fraction of its drafts, while a more accurate mechanism may incur enough proposal latency to offset its acceptance gains. Selecting an appropriate mechanism and depth requires accounting for the progress and cost of verification. This motivates our central research question:
\begin{quote}
\centering
\textit{Can speculative decoding adapt its proposal mechanism and depth to maximize verified progress per unit cost as decoding evolves?}
\end{quote}

We address this question with \apex, a learned controller that combines request-level expert selection with block-level depth adaptation. At the beginning of a request, \router selects a proposal mechanism and an initial draft depth using the available request information. The selected mechanism remains fixed for that request, while \depth adjusts its draft length at each subsequent verification block. The depth controller uses causal features describing the current prefix, generation progress, and recent verifier outcomes. This separation accommodates the different timescales of the two decisions: routing determines the speculative execution path, while depth control responds to changes in acceptance and runtime within that path.

The central modeling idea is to treat accepted draft length as censored survival feedback. When a draft is rejected at a particular position, the observation identifies how far it survived verification. When every proposed token is accepted, the observation establishes survival only through the chosen depth; it does not reveal where a longer draft would first be rejected. \apex captures this structure through position-wise rejection hazards, from which it derives survival probabilities and expected accepted length. Runtime prediction complements this model because accepting more tokens does not necessarily produce greater speedup when drafting or verification costs also increase.

\subsection{Contributions}
Our work makes the following contributions:
\begin{enumerate}

\item \textbf{Empirical characterization of speculative speed and waste.}
We examine how proposal mechanisms and draft depths behave across workloads, entropy, repetition, and execution costs. The analysis identifies conditions in which deeper speculation increases waste without proportional acceleration and motivates control using decoding-state information.

\item \textbf{Hierarchical control of heterogeneous speculators.}
We introduce \apex, which coordinates request-level selection among EAGLE-3, n-gram, and draft-model speculation with block-level depth adaptation. This structure combines mechanism selection with finer-grained responses to changing verifier feedback.

\item \textbf{A survival-based model of draft acceptance.}
We formulate accepted draft length using position-wise rejection hazards and a likelihood that handles both early rejection and fully accepted, right-censored blocks. Joint prediction of acceptance behavior and runtime provides supervision for learning the value of different depths.

\item \textbf{Trace-supervised learning of action utility.}
We construct utility targets from observed throughput, accepted progress, draft-token waste, and regret relative to observed alternatives. Multi-task prediction and candidate-ranking losses train the controller to select depths using features available at inference time.

\item \textbf{System integration and detailed evaluation.}
We integrate \apex into vLLM and evaluate it with Qwen3-8B across six workloads, including component ablations and candidate-depth sensitivity analysis. \apex-S achieves up to $5.24\times$ wall-clock speedup over autoregressive decoding and an aggregate speedup of $4.27\times$. \apex-B achieves an aggregate speedup of $3.27\times$ while reducing the mean wasted-token percentage from $79.90\%$ to $47.17\%$ relative to fixed n-gram speculation at $k=16$, a relative reduction of $41.0\%$. 
\end{enumerate}

\section{Background and Related Work}

\subsection{Autoregressive Decoding}
\label{autoreg}

Let \(p_\theta\) denote the target language model. Given a prompt or prefix \(x_{\le t}=(x_1,\ldots,x_t)\), autoregressive decoding\cite{schuurmans2024autoregressivelargelanguagemodels} generates one token at a time according to $x_{t+1} \sim p_\theta(\cdot \mid x_{\le t})$. The newly generated token is appended to the prefix and the process repeats until a stopping condition is reached. This decoding procedure preserves the target model's generation semantics, but it introduces a strict sequential dependency: producing \(T\) output tokens requires \(T\) target-model decoding steps. For large models, each step requires loading and applying the target network to the current prefix, making decoding latency scale directly with output length.

This sequential structure is especially costly for long-form generation settings such as code synthesis, multi-step reasoning, long-context completion, and software engineering tasks. The goal of inference acceleration is therefore not merely to reduce computation, but to reduce the number or cost of sequential target-model invocations while preserving the output distribution or verification semantics of the target model.

SD accelerates autoregressive generation by separating token proposal from target-model verification \cite{xia-etal-2023-speculative,leviathan2023fastinferencetransformersspeculative,chen2023acceleratinglargelanguagemodel}. A cheaper proposal mechanism \(q\) first drafts a block of \(k\) candidate tokens, $\hat{x}_{t+1:t+k} \sim q(\cdot \mid x_{\le t})$, and the target model verifies the proposed block in parallel. If multiple draft tokens are accepted, generation advances by more than one token using a single target-model verification step.

The efficiency of this procedure depends on how many proposed tokens survive verification. Let \(L_t\in\{0,\ldots,k\}\) denote the accepted draft length at block \(t\), and let \(r_t\) denote the first rejection position, with the event \(r_t>k\) corresponding to full acceptance of the draft block. If rejection occurs early, all remaining draft tokens after the first rejected token are discarded. Thus, the useful progress of a speculative block is not the proposed depth \(k\), but the accepted length \(L_t\).

This observation makes SD a cost-sensitive procedure. A larger \(k\) increases the maximum possible progress per verifier call, but it also increases the amount of draft work that can be wasted after early rejection. The relevant quantity for efficient serving is therefore accepted progress per unit cost:
\begin{equation}
    \mathcal{A}_t(m,k) = \frac{L_t}{C_t},
\end{equation}

where \(m\) denotes the speculative mechanism and \(C_t\) denotes the block-level proposal and verification cost. A speculative configuration is profitable only when the accepted progress outweighs the extra drafting and verification overhead.

\subsection{Speculative Mechanisms}

Early SD methods use an auxiliary draft model to propose tokens for target-model verification \citep{xia-etal-2023-speculative,leviathan2023fastinferencetransformersspeculative}. Followup works improve drafting through knowledge distillation, specialized architectures, or objectives aligned with target-model acceptance \citep{bhansali2025draftverifyimprovetrainingaware,zou2026variationalspeculativedecodingrethinking}. 
Improving an individual proposal mechanism does not by itself determine when that mechanism is preferable to alternatives with different proposal costs and acceptance behavior. A fixed choice can therefore retain a costly or poorly matched speculative path even when another mechanism offers a more favorable speed--waste trade-off.

A second line of work reduces reliance on separate draft models. Medusa adds decoding heads to the target model for parallel verification of tree-structured continuations \citep{cai2024medusasimplellminference}, while SpecInfer constructs token trees for parallel candidate verification \citep{Miao_2024}. EAGLE-style methods instead use target-model hidden states to build lightweight neural speculators \citep{li2025eaglespeculativesamplingrequires,li2025eagle3scalinginferenceacceleration}. Their gains, however, still depend on proposal survival and cost.

A third class uses retrieval or surface repetition instead of neural drafting. N-gram and suffix-based methods copy continuations from the prompt, generated prefix, or external text, making them cheap and effective in repetitive regimes but brittle on novel continuations \citep{hu2024samdecodingspeculativedecoding}. This creates a complementary tradeoff: neural speculators can better model semantic continuation, while retrieval-based speculators can be faster when repeated structure is available.

\apex is orthogonal to these mechanisms; it treats mechanisms like EAGLE-3, n-gram, and draft-model speculation as candidate experts and learns which of them to use and how deeply to speculate in each decoding state. \apex can thus choose among mechanisms with different proposal costs and acceptance behavior while retaining their existing proposal and verification algorithms.

\subsection{Adaptive Speculative Decoding}

Dynamic Speculation Lookahead shows that using the same speculation lookahead for all decoding iterations is suboptimal and proposes dynamically selecting the number of draft tokens during generation \citep{mamou2024dynamicspeculationlookaheadaccelerates}. AdaEAGLE~\citep{zhang2024adaeagleoptimizingspeculativedecoding}explicitly models adaptive draft structures and uses a lightweight draft length predictor to choose the number of draft tokens for EAGLE-style speculation. BanditSpec~\citep{hou2025banditspecadaptivespeculativedecoding} formulates speculative hyperparameter selection as a multi-armed bandit problem and adapts the configuration online without training a controller. DSDE~\citep{yang2025dsdedynamicspeculativedecoding} studies dynamic speculation length in serving settings using KLD-based stability signals and adaptive caps. Production systems such as vLLM also support dynamic SD through manually specified schedules that map concurrency or batch-size ranges to a chosen number of speculative tokens \citep{kwon2023efficientmemorymanagementlarge}.

These studies show that a single fixed depth ($k$) can limit inference acceleration as acceptance behavior and serving conditions change. They motivate a broader control problem: selecting a proposal mechanism together with depths that balance verified progress, runtime, and draft-token waste. \apex addresses this problem through request-level expert selection and block-level depth adaptation, using learned rejection hazards and runtime signals to score candidate depths.

 \begin{figure*}
    
        \centering
        \includegraphics[width=0.85\linewidth]{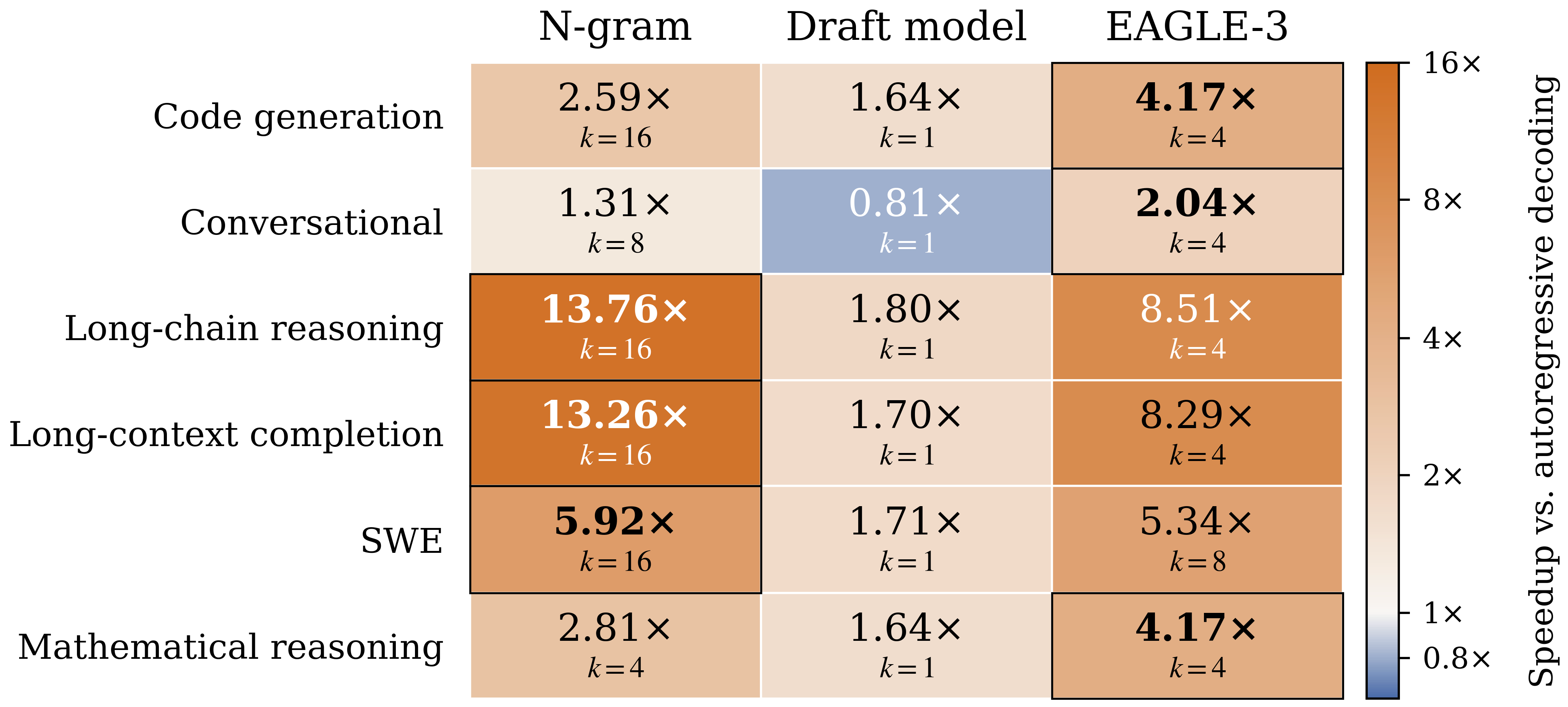}
    \caption{
    \textbf{Speedup over autoregressive decoding} across six workloads and three speculative mechanisms. Each cell reports highest speedup and its depth $k$; boxes mark the best method per workload. }
    \label{fig:depth}
\end{figure*}
\subsection{Speculative Decoding}
\label{speculative decoding}

\begin{figure*}
    
    \centering
    \begin{subfigure}{0.45\linewidth}
        \centering
        \includegraphics[width=\linewidth]{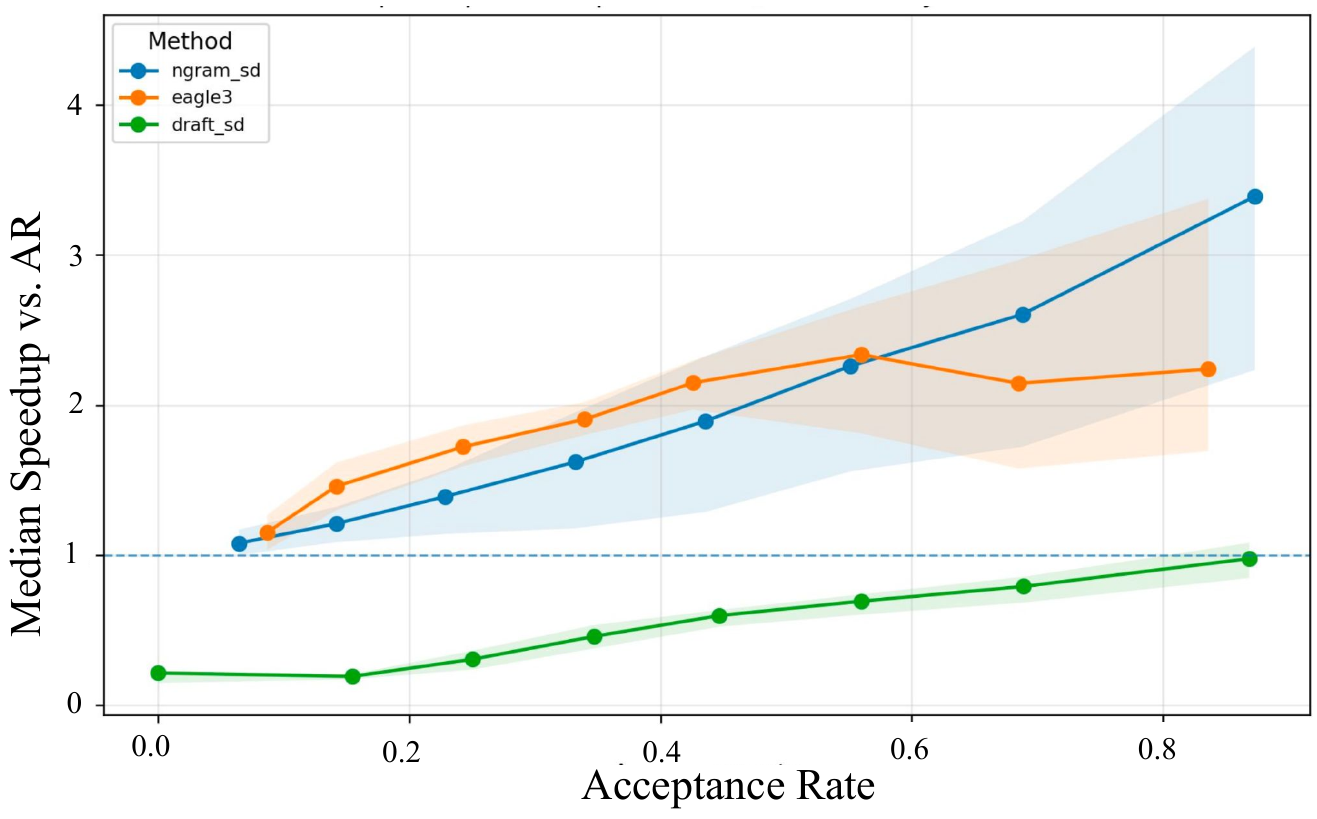}
        \caption{Acceptance and speedup}
        \label{fig:seepupacc}
    \end{subfigure}
    \begin{subfigure}{0.45\linewidth}
        \centering
        \includegraphics[width=\linewidth]{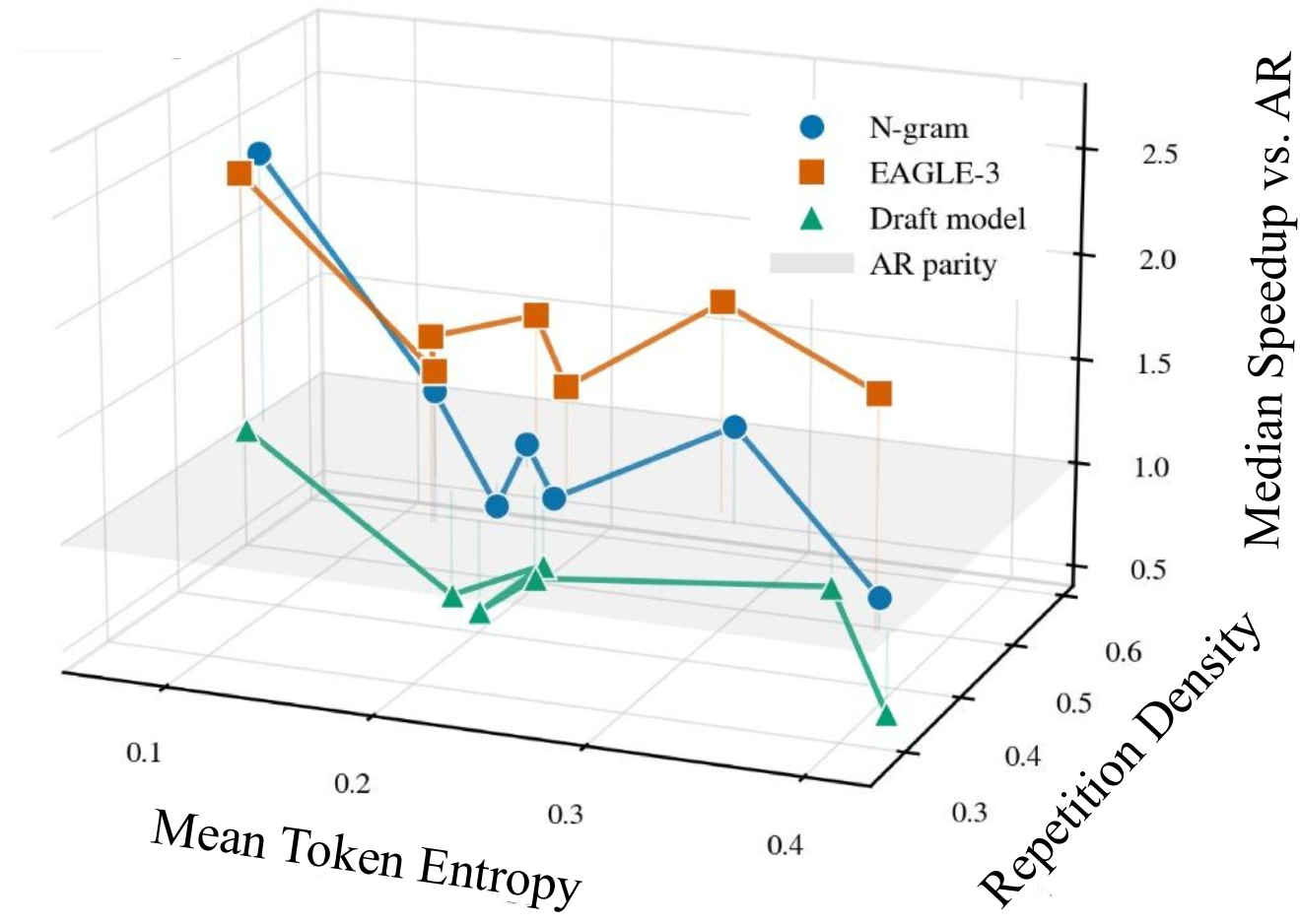}
        \caption{Entropy--repetition vs speedup}
        \label{fig:repitition}
    \end{subfigure}

    \caption{\textbf{Online signals associated with SD performance}  (a) Median speedup over autoregressive decoding versus acceptance rate, with shaded variability. (b) Median speedup versus token entropy and repetition density; the gray plane marks autoregressive performance. }
\end{figure*}

\section{Empirical Motivation: Fixed Speculation Is Not Enough}

We examine how the proposal mechanism and draft depth affect acceleration and draft-token waste. Using Qwen3-8B in vLLM, we compare autoregressive decoding with EAGLE-3, n-gram, and draft-model speculation across code generation, mathematical reasoning, long-context completion, long-chain reasoning, conversational generation, and SWE-style tasks. Workload details and full evaluation results appear in Section~\ref{sec:results}.

\paragraph{Speculative performance depends on the workload and mechanism.}
Figure~\ref{fig:depth} summarizes the highest observed speedups in the characterization set, illustrating workload-dependent differences between mechanisms. These peak results are distinct from the aggregate averages reported on a separate evaluation set in Table~\ref{tab:overall-results}. Figure~\ref{fig:seepupacc} further shows that acceptance alone does not determine acceleration: draft-model speculation remains near or below autoregressive performance in the plotted median results, even at relatively high acceptance rates, while the other mechanisms achieve substantial speedup. This motivates considering execution cost alongside acceptance when selecting a speculative mechanism.

\paragraph{Greater depth does not guarantee a better speed--waste trade-off.}
The aggregate sweep in Table~\ref{tab:overall-results} shows contrasting responses to deeper speculation. Increasing n-gram depth from $k=8$ to $k=16$ raises mean speedup from $3.59\times$ to $4.22\times$, while mean waste increases from $63.56\%$ to $79.90\%$. For EAGLE-3, increasing depth from $k=4$ to $k=16$ reduces mean speedup from $2.35\times$ to $1.37\times$ and raises mean waste from $66.25\%$ to $97.67\%$. Deeper drafting therefore offers a modest speed gain at substantially higher waste in one case and worsens both metrics in the other. These responses motivate depth selection that accounts for the active mechanism and expected verification outcome.

\paragraph{Observable decoding signals can inform control.}
Figure~\ref{fig:repitition} relates median speedup to token entropy and repetition density. Performance varies across these conditions, and the patterns differ between mechanisms. These associations motivate using entropy and repetition as controller inputs alongside recent verifier feedback. Prefix statistics describe the context in which a draft is proposed, while verifier history records how well recent drafts have performed. Both can be obtained from information available during decoding, allowing adaptation without requiring workload labels at inference time.

\paragraph{Verifier feedback reveals where drafts fail.}
For a draft of length $k$, first rejection at position $r$ leaves $r-1$ accepted draft tokens and $k-r+1$ discarded draft tokens. Full acceptance provides a different observation: the draft survives through depth $k$, but the outcome of extending it further remains unknown. Verifier feedback thus contains both observed rejection events and right-censored full-acceptance events. This structure motivates modeling position-wise rejection hazards, which describe how acceptance can change with depth and provide supervision for adaptive speculative control.

\section{APEX: Breathing Adaptivity into Static Speculation}

We propose \apex, a learned online controller that selects a proposal mechanism and initial depth at request granularity and adjusts speculation depth at block granularity. Its learned utility balances throughput and accepted progress against discarded draft tokens. \apex operates at two coupled timescales. \textbf{\router} selects a speculative expert (EAGLE-3, n-gram, or draft-model speculation) and its initial depth for each request, while \textbf{\depth} adapts depth $k$ at the block level. This separation reflects systems constraints: expert selection affects model placement, batching, and worker routing, whereas depth can change more frequently within the selected path.

\begin{figure*}[t]
    \centering
    \includegraphics[width=\linewidth]{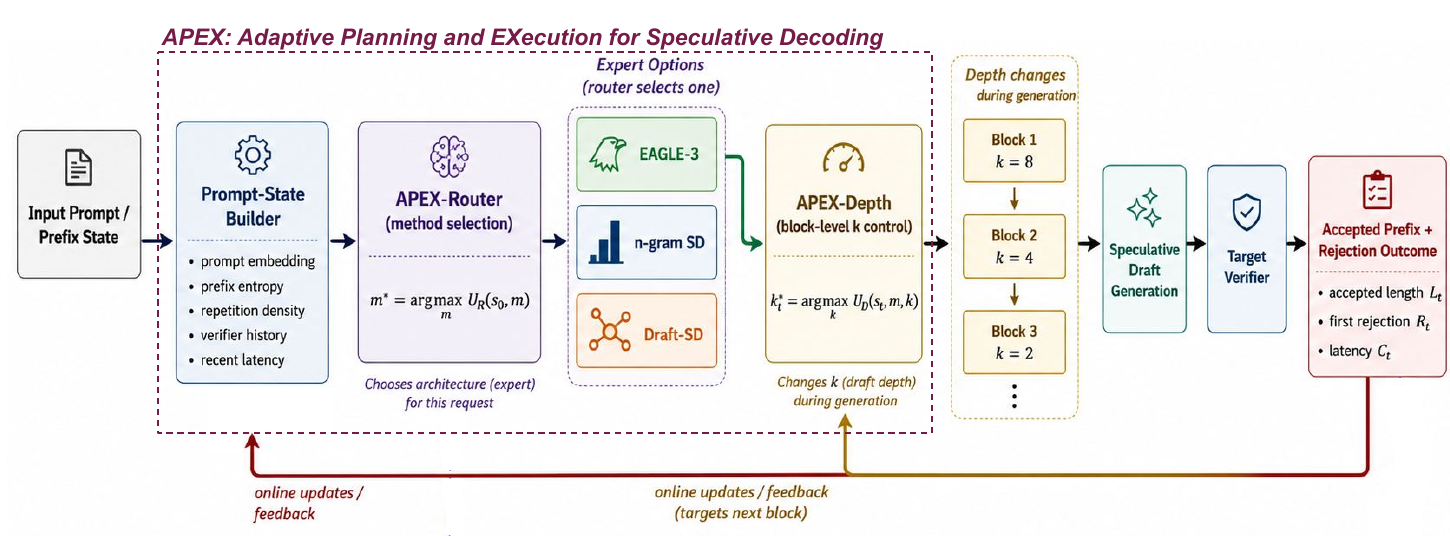}
    \caption{\textbf{\apex overview.} The router selects a speculative expert and initial depth from prompt features. After each verifier block, the depth controller uses the updated decoding state and prior verifier feedback to select the next depth. Observed acceptance and runtime update the controller state; network parameters remain fixed during inference.}
    \label{fig:proposed_flow}
\end{figure*}

Figure~\ref{fig:proposed_flow} summarizes the framework. We first define the decoding state and request-level routing decision, then formulate accepted draft length as censored survival feedback. This formulation provides the acceptance model used by \depth, which learns acceptance, runtime, and utility predictions from a shared representation.

\subsection{Problem Formulation and State Representation}
\label{sec:problem-formulation}
\label{app:state}

Let $p_\theta$ denote the target language model and let $m\in\mathcal{M}$ denote a proposal mechanism, where
\begin{equation}
    \mathcal{M} = \{\text{EAGLE-3},\text{n-gram SD},\text{Draft-SD}\}.
\end{equation}
We index verifier blocks by $t$ and denote the prefix available before block $t$ by $c_t$. Given this prefix, mechanism $m$ proposes a draft block $d_t$ of length $k_t$ according to $q_m(\cdot\mid c_t;k_t)$, its induced proposal distribution. The target model then verifies the draft block. Let $L_t\in\{0,\ldots,k_t\}$ be the number of accepted draft tokens. If $L_t=k_t$, the full block is accepted; if $L_t<k_t$, the verifier rejects at position $L_t+1$. We define the discarded draft-token count and waste fraction as
\begin{equation}
R_t=k_t-L_t, \qquad W_t=\frac{R_t}{k_t}.
\end{equation}
Here, $R_t$ includes the first rejected token and the discarded suffix, and $W_t$ is the fraction of drafted tokens that are not accepted. Since $W_t$ measures token utilization rather than wasted computation, we measure latency separately to assess the speed--waste trade-off.

Let $C_t$ denote block-$t$ wall-clock latency, including drafting and verification, and $\mathrm{TPS}_t$ the executed action's throughput. \apex uses these observations to learn a block-level utility balancing accepted progress, runtime, and speculative waste. In our experiments, $K_{\max}=16$, and candidate depths form a finite set $\mathcal{K}\subseteq\{1,\ldots,K_{\max}\}$.

\paragraph{\textbf{Request-level state}}
The request-level state $s_0$ contains the information available before generation begins:
\begin{equation}
    \begin{aligned}
s_0=[&e_{\mathrm{prompt}},\ell_{\mathrm{prompt}}, \rho_{\mathrm{prompt}}, \eta_{\mathrm{prompt}},c_{\mathrm{decode}}].
\end{aligned}
\end{equation}
Here, $e_{\mathrm{prompt}}$ is a fixed-dimensional prompt embedding, $\ell_{\mathrm{prompt}}$ is the prompt length, $\rho_{\mathrm{prompt}}$ contains repetition and lexical-structure features, $\eta_{\mathrm{prompt}}$ contains entropy or token-distribution features, and $c_{\mathrm{decode}}$ contains decoding configuration variables such as temperature and maximum generation length.

\paragraph{\textbf{Block-level state}}
The block-level state $s_t$ contains information available before verifier block $t$:
\begin{equation}
    \begin{aligned}
s_t=[&e_{\mathrm{prompt}},\ell_{\mathrm{prompt}}, c_{\mathrm{decode}}, \pi_t,\mathbf{h}_{t-1},\mathbf{g}_t].
\end{aligned}
\end{equation}
Here, $\pi_t$ denotes generation progress, $\mathbf{h}_{t-1}$ summarizes verifier feedback from blocks $1,\ldots,t-1$, and $\mathbf{g}_t$ contains statistics computed from the current prefix $c_t$. We supply the active mechanism $m$ and candidate depth $k$ separately when constructing state-action features.

The history vector $\mathbf{h}_{t-1}$ includes the number of observed verifier blocks, rolling mean and variance of accepted length, first-rejection position statistics, full-acceptance rate, rejection rate, mean rejected-token count, and short-window summaries over the most recent verifier blocks. The prefix vector $\mathbf{g}_t$ includes entropy and repetition features. These features are causal: the current block outcome is never used to predict the current block.

\subsection{APEX-Router}
\label{sec:apex-router}

\router jointly selects the speculative expert and initial depth expected to provide the best speed--cost trade-off for each request before generation. It scores initial actions $a=(m,k_0)$, where $m\in\mathcal{M}$ is the speculative mechanism and $k_0$ is the initial depth. Let $\mathcal{A}_R\subseteq\mathcal{M}\times\mathcal{K}$ denote the admissible initial actions. Using the pre-generation state $s_0$, it predicts
\begin{equation}
\widehat{U}_R(s_0,m,k_0) = f_{\theta_R}\!\left(\phi_R(s_0,m,k_0)\right),
\end{equation}
where $\phi_R$ combines the request features with the candidate method and initial depth, and $f_{\theta_R}$ is a neural scoring function. The selected action is
\begin{equation}
(m^\star,k_0^\star) =\arg\max_{(m,k_0)\in\mathcal{A}_R} \widehat{U}_R(s_0,m,k_0).
\end{equation}

The router is trained from completed request-level traces rather than workload labels. Each observed request--method--initial-depth combination supplies a training example:
\[
(s_0^{(r)},m,k_0) \longrightarrow u_R(r,m,k_0),
\]
where $s_0^{(r)}$ is the initial state of request $r$ and $u_R(r,m,k_0)$ is its empirical request-level utility label. The observed label $u_R$ provides offline supervision for the predicted score $\widehat{U}_R$. The resulting acceptance, latency, and waste are execution outcomes; they are not inputs available when routing a new request.

At inference, the router scores each admissible method--initial-depth pair using only $s_0$ and the candidate action. The selected action determines the speculative decoding path and initializes its depth. The mechanism $m^\star$ remains fixed for the request, the first block uses $k_1=k_0^\star$, and \depth controls subsequent depths.

Once the expert is selected, the depth decision depends on how its drafts survive verification as generation evolves. We next formulate this acceptance feedback before describing the depth controller.

\subsection{Acceptance as Censored Survival Feedback}
\label{sec:acceptance-survival}
\label{app:survival}

The accepted length of a speculative block contains more information than an aggregate acceptance rate. A partially accepted block identifies the first rejected position, while a fully accepted block establishes only that rejection did not occur within the executed depth. \apex models these two outcomes using a discrete survival formulation.

For a candidate depth $k$, define $J_t$ as the first rejected draft position using one-indexed positions. Thus, $J_t=j$ means the first $j-1$ draft tokens were accepted and the $j$-th draft token was rejected. If the entire block is accepted, then $J_t>k$. The accepted length is
\begin{equation}
    L_t= \begin{cases} J_t-1, & J_t\leq k,\\
k, & J_t>k.
\end{cases}
\end{equation}

\paragraph{Position-wise rejection hazards.}
For each position $j\in\{1,\ldots,K_{\max}\}$, the model predicts a discrete hazard:
\begin{equation}
\label{eq:rejection-hazard}
\hat{q}_{t,j}(k) =P_{\theta_D}(J_t=j\mid J_t\geq j,s_t,m,k),
\end{equation}
where $\theta_D$ denotes the parameters of the depth controller. This is the conditional probability of rejection at position $j$, given that all preceding draft tokens have been accepted.

The probability that the first $j$ draft tokens survive is
\begin{equation}
\label{eq:draft-survival}
\begin{aligned}
\hat{S}_{t,j}(k) =P_{\theta_D}(J_t>j\mid s_t,m,k)=\prod_{i=1}^{j}\left(1-\hat{q}_{t,i}(k)\right).
\end{aligned}
\end{equation}
The expected accepted length follows as
\begin{equation}
\label{eq:expected-accepted-length}
\widehat{\mathbb{E}}[L_t\mid s_t,m,k] =\sum_{j=1}^{k}\hat{S}_{t,j}(k),
\end{equation}
using the identity $\mathbb{E}[L]=\sum_{j=1}^{k}P(L\geq j)$.

\paragraph{Accepted-length likelihood.}
If the observed accepted length is $L_t=\ell<k$, the first rejected position is $J_t=\ell+1$. The likelihood is therefore
\begin{equation}
    P_{\theta_D}(L_t=\ell\mid s_t,m,k) = \left[ \prod_{j=1}^{\ell}(1-\hat{q}_{t,j}(k)) \right]\hat{q}_{t,\ell+1}(k).
\end{equation}
The empty product equals one when $\ell=0$. If $L_t=k$, the entire block is accepted and the rejection position is right-censored beyond $k$:
\begin{equation}
    P_{\theta_D}(L_t=k\mid s_t,m,k) =\prod_{j=1}^{k}(1-\hat{q}_{t,j}(k)).
\end{equation}
This observation contributes evidence of survival through position $k$ without imposing a rejection at position $k+1$.

The survival loss over $N$ training blocks is
\begin{equation}
\label{eq:survival-loss}
\mathcal{L}_{\mathrm{surv}} =-\frac{1}{N}\sum_{t=1}^{N} \log P_{\theta_D}(L_t\mid s_t,m_t,k_t).
\end{equation}
This objective trains the controller to capture the first-rejection structure of SD. Acceptance prediction is then combined with runtime and utility supervision, since accepted length alone does not determine the efficiency of a candidate depth.

\subsection{APEX-Depth}
\label{sec:apex-depth}

\depth adapts speculation depth to balance accepted progress, latency, and draft-token waste. Its operation begins with the speculative method and initial depth selected by \router. Before each subsequent block, it observes the causal state $s_t$, evaluates the candidate depths, and selects the depth with the highest predicted utility.

\paragraph{\textbf{Shared representation}}
For each candidate depth $k\in\mathcal{K}$, the controller constructs
\begin{equation}
z_t(k)=\phi_D(s_t,m^\star,k), \qquad v_t(k)=g_{\theta_D}(z_t(k)).
\end{equation}
Here, $\phi_D$ standardizes numerical features, embeds categorical variables, and appends the prompt representation. The shared encoder therefore evaluates each depth in the context of the current generation state.

Four prediction heads receive $v_t(k)$. Each head is a task-specific output branch of the same network: the survival, cost, and throughput heads predict execution outcomes, while the utility head predicts the action score used for depth selection.

\paragraph{\textbf{Survival head}}
The survival head outputs a vector of $K_{\max}$ conditional rejection probabilities:
\begin{equation}
    \hat{\mathbf{q}}_t(k) =f_{\mathrm{surv}}(v_t(k)).
\end{equation}
Its $j$-th entry is the hazard $\hat{q}_{t,j}(k)$ defined in Equation~\eqref{eq:rejection-hazard}. The first $k$ entries determine the survival curve and expected accepted length through Equations~\eqref{eq:draft-survival}--\eqref{eq:expected-accepted-length}. This head is trained with the censored likelihood in Section~\ref{sec:acceptance-survival}, allowing fully accepted and partially rejected blocks to provide appropriate supervision.

\paragraph{\textbf{Cost head}}
The cost head predicts the wall-clock cost of executing the candidate block, including drafting and verification. Its scalar output is the predicted log latency:
\begin{equation}
    \hat{c}_t(k) =f_{\mathrm{cost}}(v_t(k)) =\widehat{\log C_t(k)}.
\end{equation}
This head provides runtime supervision because a candidate depth can accept many tokens but still be slow when proposal or verification overhead is high.

\paragraph{\textbf{Throughput head}}
The throughput head predicts the rate of output progress achieved by the candidate action. Its scalar output is
\begin{equation}
    \hat{\tau}_t(k) =f_{\mathrm{tps}}(v_t(k)) =\widehat{\log(1+\mathrm{TPS}_t(k))}.
\end{equation}
Where the cost head models elapsed time, the throughput head models the observed progress rate. Their targets are obtained from measured execution traces, and their regression losses are defined in Section~\ref{app:losses}.

\paragraph{\textbf{Utility head}}
The utility head predicts the final learned action value:
\begin{equation}
    \widehat{U}_D(s_t,m^\star,k) =f_{\mathrm{util}}(v_t(k)).
\end{equation}
The hatted quantity $\widehat{U}_D$ is a network prediction made before the candidate block executes. During training, it is fitted to an observed scalar label $u_{D,t}$ computed from execution outcomes, as defined in Equation~\eqref{equtility}. The utility-regression loss in Equation~\eqref{eq:depth-utility-loss} directly connects this prediction to its label, while the ranking loss supervises candidate ordering. The other three heads ground the shared representation in measurable outcomes.

\paragraph{\textbf{Select, execute, and update}}
The controller chooses the highest-scoring candidate:
\begin{equation}
\label{ktstar}
k_t^\star =\arg\max_{k\in\mathcal{K}} \widehat{U}_D(s_t,m^\star,k), \qquad t\geq 2.
\end{equation}
The utility head supplies the deployed selection score; the other heads support it through shared training. The chosen speculative block is then executed, and its observed outcomes update the state for the next decision. This allows depth to respond to changes in acceptance behavior and execution cost throughout generation.

\subsection{Trace-Derived Training Targets}
\label{sec:method-targets}
\label{app:regret}
\label{app:block-utility}

We train \apex from offline collected speculative decoding traces. Each depth-controller training example corresponds to an observed verifier block under a particular state-action pair:
\[
(s_t,m_t,k_t) \longrightarrow (L_t,R_t,W_t,C_t,\mathrm{TPS}_t).
\]
The input contains only information available before executing the block, while the targets contain the outcomes observed after verification. Accepted length supervises the survival head, latency and throughput supervise the runtime heads, and the following utility target supervises the utility head. We use lowercase $u$ for observed utility labels and hatted $\widehat{U}$ for their learned predictions.

\paragraph{\textbf{Dataset-derived regret targets}}
The trace dataset provides offline comparisons among speculative configurations. For a request $r$, let $\mathcal{D}(r)$ denote the set of method-depth actions observed in the dataset, and let $\mathrm{TPS}_{r,m,k}$ denote the recorded request-level throughput under action $(m,k)$. We define the best observed global throughput as
\begin{equation}
    B_{\mathrm{global}}(r) =\max_{(m,k)\in\mathcal{D}(r)}\mathrm{TPS}_{r,m,k},
\end{equation}
and the best observed throughput within a fixed mechanism as
\begin{equation}
    B_{\mathrm{method}}(r,m) =\max_{k:(m,k)\in\mathcal{D}(r)} \mathrm{TPS}_{r,m,k}.
\end{equation}
For a training block $t$ belonging to request $r_t$, the global and same-method regret terms are
\begin{equation}
    G_t =\frac{[B_{\mathrm{global}}(r_t)-\mathrm{TPS}_t]_+} {B_{\mathrm{global}}(r_t)},
\end{equation}
\begin{equation}
    M_t =\frac{[B_{\mathrm{method}}(r_t,m_t)-\mathrm{TPS}_t]_+} {B_{\mathrm{method}}(r_t,m_t)},
\end{equation}
where $[x]_+=\max(x,0)$. These normalized penalties compare block throughput with the best observed request-level reference throughputs. The global term accounts for alternatives across mechanisms and depths, while the same-method term accounts for depth choices within the active mechanism. Both terms are used only to construct training targets and are not required during inference.

\paragraph{\textbf{Block-level utility}}
For each observed training block, the depth-utility label is
\begin{equation}
\label{equtility}
\begin{aligned}
u_{D,t}={}&\log(\mathrm{TPS}_t) +\alpha_A\log(L_t+\epsilon)-\alpha_W W_t-\alpha_G G_t-\alpha_M M_t,
\end{aligned}
\end{equation}
where $\epsilon=0.1$ keeps the accepted-length term finite when $L_t=0$. This is the trace-derived label for the executed action $(s_t,m_t,k_t)$; $\widehat{U}_D(s_t,m_t,k_t)$ is the network's prediction of that label.

The utility combines speed, accepted progress, waste, and offline regret. High accepted length is useful only if the resulting progress justifies its runtime, while low waste is useful only if the system still makes enough progress per verifier call. The coefficients allow different training preferences for this speed--waste trade-off.

\subsection{Training Objective for APEX-Depth}
\label{app:losses}

The \depth network is multi-task by design. The survival loss in Equation~\eqref{eq:survival-loss} supervises the rejection hazards. The cost and throughput losses supervise runtime predictions, while utility regression and candidate ranking train the score used for depth selection.

\paragraph{\textbf{Cost and throughput losses}}
Using the observed action $(s_t,m_t,k_t)$ for each training block, we train the cost head with
\begin{equation}
    \mathcal{L}_{\mathrm{cost}} =\frac{1}{N}\sum_{t=1}^{N} \rho\!\left(\hat{c}_t(k_t)-\log C_t\right),
\end{equation}
where $\rho(\cdot)$ is the SmoothL1 loss. The throughput head is trained with
\begin{equation}
    \mathcal{L}_{\mathrm{tps}} =\frac{1}{N}\sum_{t=1}^{N} \rho\!\left( \hat{\tau}_t(k_t)-\log(1+\mathrm{TPS}_t) \right).
\end{equation}

\paragraph{\textbf{Utility regression}}
The utility head is trained by comparing its prediction for each observed state-action pair with the corresponding trace-derived label:
\begin{equation}
\label{eq:depth-utility-loss}
\mathcal{L}_{\mathrm{util}} =\frac{1}{N}\sum_{t=1}^{N} \rho\!\left(\widehat{U}_D(s_t,m_t,k_t)-u_{D,t}\right).
\end{equation}
Thus, the same predicted score used to select depths at inference is directly supervised during training.

\paragraph{\textbf{Candidate ranking}}
Deployment depends on the ordering of candidate depths, as well as the accuracy of their predicted utilities. We therefore group observed candidate rows by prompt, speculative method, and block index. For a group $g$, let $\mathcal{K}_g$ denote the depths present, and let $u_{D,g}(k)$ denote the observed utility label at depth $k$. Writing $s_{g,k}$ for that row's pre-block state and $m_g$ for the group's method, its predicted score is
\begin{equation}
    \widehat{U}_{D,g}(k) =\widehat{U}_D(s_{g,k},m_g,k).
\end{equation}
The preferred depth is
\begin{equation}
    k_g^+=\arg\max_{k\in\mathcal{K}_g}u_{D,g}(k).
\end{equation}

The ranking loss is
\begin{equation}
    \mathcal{L}_{\mathrm{rank}} =-\frac{1}{|\mathcal{G}|} \sum_{g\in\mathcal{G}} \log \frac{\exp(\widehat{U}_{D,g}(k_g^+))} {\sum_{k\in\mathcal{K}_g}\exp(\widehat{U}_{D,g}(k))},
\end{equation}

where $\mathcal{G}$ is the set of rank groups. This term encourages the depth with the highest observed training utility in each group to receive the highest predicted score.

\paragraph{\textbf{Combined objective}}
The depth controller is trained with
\begin{equation}
    \begin{aligned}
\mathcal{L}_D={}&
\lambda_s\mathcal{L}_{\mathrm{surv}} +\lambda_c\mathcal{L}_{\mathrm{cost}} +\lambda_v\mathcal{L}_{\mathrm{tps}}+\lambda_u\mathcal{L}_{\mathrm{util}} +\lambda_r\mathcal{L}_{\mathrm{rank}}.
\end{aligned}
\end{equation}

In our implementation, the default loss weights are
\begin{equation}
    \begin{aligned}
(\lambda_s,\lambda_c,\lambda_v,\lambda_u,\lambda_r)=(1.00,0.20,0.30,1.00,1.00).
\end{aligned}
\end{equation}

\subsection{APEX Variants}
\label{sec:apex-variants}

We train three \apex variants with identical architectures, features, candidate-depth interfaces, data, and deployment procedures; only the utility weights in Equation~\eqref{equtility} differ, producing distinct speed--waste operating points.

\textsc{APEX-S} (Speed) uses $(\alpha_A,\alpha_W,\alpha_G,\alpha_M)=(0.20,0.20,0.20,0.15)$ and places greater relative emphasis on throughput competitiveness. The default \textsc{APEX-B} (Balanced) uses $(0.30,0.35,0.15,0.20)$ to balance throughput, accepted progress, and waste. \textsc{APEX-E} (Efficient) uses $(0.40,0.50,0.10,0.25)$ and assigns a larger penalty to waste. These weights specify training preferences; they do not guarantee that every profile is Pareto-efficient or that \textsc{APEX-E} always has the lowest waste. Additional ablations are reported in Section~\ref{ksweep}.

\subsection{Online Deployment}
\label{sec:apex-deployment}

At inference, the router selects the speculative expert $m^\star$ and initial depth $k_0^\star$. After each verifier block, \apex updates its causal state with observed outcomes, scores every candidate depth in $\mathcal{K}$, and selects the next block's depth. The controller adapts its actions using the evolving state; its network parameters remain fixed during generation.

Algorithm~\ref{alg:apex-online} summarizes this procedure. Here, $\operatorname{SpeculateAndVerify}$ drafts a block using the selected mechanism, applies target-model verification, and records the execution outcomes. It returns the emitted output $\Delta y_t$ and observations $o_t=(L_t,R_t,W_t,C_t,\mathrm{TPS}_t)$. The output extends the prefix, while the observations update the history used for the next decision.

\begin{algorithm}[t]
\caption{\apex Online Speculative Decoding}
\label{alg:apex-online}
\begin{algorithmic}[1]
\REQUIRE Prompt $x$, decoding configuration $c_{\mathrm{decode}}$, target model $p_\theta$, initial actions $\mathcal{A}_R$, depths $\mathcal{K}$
\STATE $s_0\leftarrow\operatorname{RequestState}(x,c_{\mathrm{decode}})$
\STATE $(m^\star,k_0^\star)\leftarrow
    \displaystyle\arg\max_{(m,k_0)\in\mathcal{A}_R}
    \widehat{U}_R(s_0,m,k_0)$
\STATE $\mathbf{h}_0\leftarrow\emptyset$, $y\leftarrow\emptyset$
\FOR{$t=1,2,\ldots$ until generation completes}
    \STATE $c_t\leftarrow x\mathbin{\Vert}y$
    \STATE $s_t\leftarrow\operatorname{State}(s_0,c_t,\mathbf{h}_{t-1})$
    \IF{$t=1$}
        \STATE $k_t\leftarrow k_0^\star$
    \ELSE
        \STATE $k_t\leftarrow
            \displaystyle\arg\max_{k\in\mathcal{K}}
            \widehat{U}_D(s_t,m^\star,k)$
    \ENDIF
    \STATE $(\Delta y_t,o_t)\leftarrow
        \operatorname{SpeculateAndVerify}(p_\theta,m^\star,k_t,c_t)$
    \STATE $y\leftarrow y\mathbin{\Vert}\Delta y_t$
    \STATE $\mathbf{h}_t\leftarrow\operatorname{Update}(\mathbf{h}_{t-1},o_t)$
\ENDFOR
\STATE \textbf{return} $y$
\end{algorithmic}
\end{algorithm}

\begin{table*}[t]
\centering
\caption{\textbf{Overall decoding performance.} \apex is compared against autoregressive decoding and fixed speculative baselines. Higher speedup is better; lower wasted-token percentage is better.}
\label{tab:overall-results}
\begin{tabular}{llccc}
\toprule
\textbf{Method} & \textbf{Depth} & \textbf{Speedup vs. AR $\uparrow$}  & \textbf{Wasted Token \% $\downarrow$} & \textbf{WAS $\uparrow$} \\
\midrule
\multirow{4}{*}{EAGLE-3} & $k=2$ & 1.71 $\times$ & 64.71\% & 2.64 \\
 & $k=4$ & 2.35 $\times$ & 66.25\% & 3.55 \\
 & $k=8$ & 1.94 $\times$ & 87.11\% & 2.23 \\
 & $k=16$ & 1.37 $\times$ & 97.67\% & 1.41 \\
\midrule
\multirow{4}{*}{n-gram SD} & $k=2$ & 1.76 $\times$ & 54.56\% & 3.23 \\
 & $k=4$ & 2.79 $\times$ & 54.14\% & 5.16 \\
 & $k=8$ & 3.59 $\times$ & 63.56\% & 5.64 \\
 & $k=16$ & 4.22 $\times$ & 79.90\% & 5.28 \\
\midrule
\multirow{4}{*}{Draft-SD} & $k=2$ & 0.96 $\times$ & 27.82\% & 3.44 \\
 & $k=4$ & 0.87 $\times$ & 38.39\% & 2.26 \\
 & $k=8$ & 0.73 $\times$ & 52.46\% & 1.39 \\
 & $k=16$ & 0.50 $\times$ & 68.98\% & 0.72 \\
\midrule
\textbf{APEX-S }& adaptive & 4.27 $\times$ & 55.30\% & \textbf{7.73} \\
\textbf{APEX-B }& adaptive & 3.27 $\times$ & 47.17\% & 6.93 \\
\textbf{APEX-E }& adaptive & 1.83 $\times$ & 63.69\% & 2.87 \\
\end{tabular}
\end{table*}

\begin{figure*}[t]
    
    \centering
        \centering
        \includegraphics[width=0.95\linewidth]{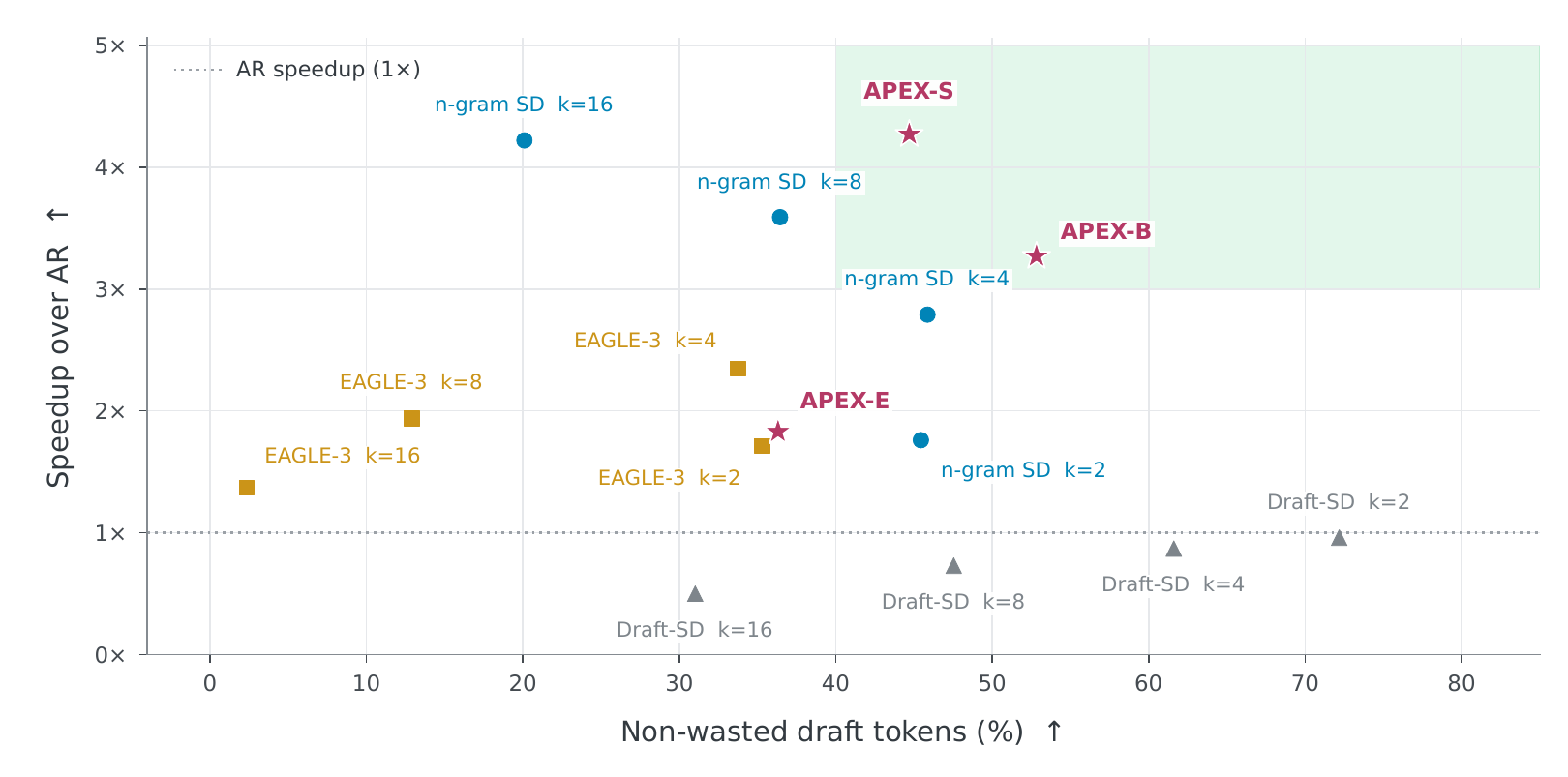}
    \caption{\textbf{Speedup vs Token Wastage.} The plot shows the trend of Speedup (vs AR) vs Token Wastage across different depths and mechanisms. The horizontal axis reports non-wasted draft tokens ($100-\mathrm{Waste}\%$), and the vertical axis reports speedup over autoregressive decoding.}
    \label{fig:trend}
\end{figure*}

\section{Results and Discussion}
\label{sec:results}

We evaluate whether \apex can maintain substantial inference acceleration while reducing discarded draft tokens relative to fixed speculative configurations. We first examine aggregate performance and workload-specific trade-offs, then study the contributions of individual controller components and the sensitivity to the candidate-depth set.

\subsection{Experimental Setup}
\label{sec:experimental-setup}

\paragraph{\textbf{Target model and serving system}}
We implement \apex in vLLM with Qwen3-8B as the target verifier. All evaluated methods use the same target model, allowing us to compare the acceleration and draft-token utilization achieved by different proposal mechanisms, fixed depths, and adaptive controller profiles.

\paragraph{\textbf{Workloads and datasets}}
We evaluate six workloads: code generation using MBPP~\citep{austin2021program} and HumanEval~\citep{chen2021evaluating}; conversational generation using UltraChat-200k~\citep{ding2023enhancingchatlanguagemodels}; long-chain reasoning using GPQA~\citep{rein2023gpqagraduatelevelgoogleproofqa}; long-context completion using The Stack Smol~\citep{Kocetkov2022TheS3}; software engineering using SWE-bench Lite~\citep{jimenez2024swebench}; and mathematical reasoning using GSM8K~\citep{cobbe2021training} and AIME-style data~\citep{maa_aime}. We follow the dataset-provided splits for training and evaluation.

\paragraph{\textbf{Baselines and operating profiles}}
We compare \apex with autoregressive decoding and fixed-depth EAGLE-3, n-gram, and draft-model speculation (Draft-SD). The fixed-depth sweep considers $k\in\{1,2,4,8,16\}$. We evaluate all three \apex operating profiles: APEX-S, APEX-B, and APEX-E.

\paragraph{\textbf{Metrics}}
We measure wall-clock latency speedup relative to autoregressive decoding as
\begin{equation}
    \mathrm{Speedup}=\frac{\mathrm{Latency}_{\mathrm{AR}}}{\mathrm{Latency}_{\mathrm{method}}}
\end{equation}
Values above one indicate lower end-to-end latency. We also report wasted draft-token percentage,
\begin{equation}
    \mathrm{Waste}_{\%}=100\times \frac{\mathrm{Rejected\ Draft\ Tokens}}{\mathrm{Total\ Draft\ Tokens}}
\end{equation}
which measures the fraction of proposed tokens discarded during speculation.

To summarize the speed--waste trade-off, we report \emph{Waste-Amortized Speedup} (WAS):
\begin{equation}
\label{waseq}
\mathrm{WAS}
=
\frac{\mathrm{Speedup}}{\mathrm{Waste}_{\%}/100}.
\end{equation}
Speedup and waste are averaged over the evaluated requests, and WAS is computed from the corresponding unrounded averages. Higher WAS is better, but we interpret it alongside latency speedup and waste to retain the distinction between these outcomes. A configuration is Pareto-efficient if no alternative is at least as fast and no more wasteful, with a strict improvement in at least one metric.

\subsection{Overall Results}
\label{sec:main-results}

Table~\ref{tab:overall-results} reports aggregate performance, and Figure~\ref{fig:trend} visualizes the same measurements as speedup versus the percentage of non-wasted draft tokens. APEX-S achieves the highest reported aggregate speedup of $4.27\times$ with $55.30\%$ waste. The fastest fixed baseline, n-gram speculation at $k=16$, reaches $4.22\times$ speedup with $79.90\%$ waste. Thus, APEX-S slightly improves the reported speedup while reducing the wasted-token percentage by $30.8\%$ relative to this baseline.

APEX-B provides a different operating point, achieving $3.27\times$ speedup with $47.17\%$ waste. Relative to n-gram speculation at $k=16$, this corresponds to a $41.0\%$ reduction in the wasted-token percentage, at the cost of lower acceleration. APEX-S and APEX-B obtain the highest aggregate WAS values of $7.73$ and $6.93$, respectively, compared with the best fixed-baseline WAS of $5.64$ from n-gram speculation at $k=8$.

The fixed baselines illustrate why speed and waste must be considered jointly. Increasing n-gram depth from $k=8$ to $k=16$ raises speedup from $3.59\times$ to $4.22\times$, but increases waste from $63.56\%$ to $79.90\%$, reducing WAS from $5.64$ to $5.28$. EAGLE-3 reaches its highest aggregate speedup at $k=4$; increasing its depth to $k=16$ reduces speedup from $2.35\times$ to $1.37\times$ while raising waste from $66.25\%$ to $97.67\%$. Conversely, Draft-SD at $k=2$ has only $27.82\%$ waste but achieves $0.96\times$ speedup. Efficient draft-token utilization alone therefore does not ensure lower end-to-end latency.

The three learned profiles also require separate evaluation. APEX-E achieves $1.83\times$ speedup with $63.69\%$ waste and is outperformed by both APEX-S and APEX-B on both metrics. Its efficiency-oriented objective does not produce the lowest aggregate waste, highlighting the importance of evaluating the realized behavior of each profile across workloads.

\begin{table*}[t]
\centering
\caption{Workload-level speed--waste comparison for \textbf{code generation}.  }
\label{tab:workload-code-generation}
\begin{tabular}{lcccc}
\toprule
\textbf{Method} & \textbf{Depth} & \textbf{Speedup vs. AR $\uparrow$}  & \textbf{Wasted Token \% $\downarrow$} & \textbf{WAS $\uparrow$}\\
\midrule
\multirow{4}{*}{EAGLE-3} & $k=2$ & 2.11$\times$ & 44.50\% & 4.74 \\
& $k=4$ & 2.78$\times$ & 55.50\% & 5.01 \\
 & $k=8$ & 2.01$\times$ & 87.38\% & 2.30 \\
 & $k=16$ & 1.27$\times$ & 98.31\% & 1.29 \\
 \midrule
 \multirow{4}{*}{n-gram SD} & $k=2$ & 2.08$\times$ & 46.00\% & 4.52 \\
 & $k=4$ & 2.31$\times$ & 67.25\% & 3.43 \\
 & $k=8$ & 3.47$\times$ & 69.13\% & 5.02 \\
 & $k=16$ & 4.25$\times$ & 79.69\% & 5.33 \\
  \midrule
\multirow{4}{*}{Draft-SD} & $k=2$ & 1.00$\times$ & 22.06\% & 4.53 \\
& $k=4$ & 0.90$\times$ & 33.74\% & 2.67 \\
 & $k=8$ & 0.71$\times$ & 51.88\% & 1.37 \\
 & $k=16$ & 0.49$\times$ & 69.30\% & 0.71 \\
  \midrule
APEX-S & adaptive & 2.80$\times$ & 76.00\% & 3.68 \\
APEX-B & adaptive & 1.99$\times$ & 56.83\% & 3.50 \\
APEX-E & adaptive & 0.99$\times$ & 75.50\% & 1.31 \\
\bottomrule
\end{tabular}
\end{table*}

\begin{table*}[t]
\centering
\caption{Workload-level speed--waste comparison for \textbf{mathematical reasoning}.  }
\label{tab:workload-mathematical-reasoning}
\begin{tabular}{lcccc}
\toprule
\textbf{Method} & \textbf{Depth} & \textbf{Speedup vs. AR $\uparrow$}  & \textbf{Wasted Token \% $\downarrow$} & \textbf{WAS $\uparrow$}\\
\midrule
\multirow{4}{*}{EAGLE-3} & $k=2$ & 1.55$\times$ & 72.50\% & 2.14 \\
& $k=4$ & 2.50$\times$ & 62.50\% & 4.00 \\
& $k=8$ & 3.22$\times$ & 72.25\% & 4.46 \\
& $k=16$ & 2.05$\times$ & 93.44\% & 2.19 \\
\midrule
 \multirow{4}{*}{n-gram SD} & $k=2$ & 1.89$\times$ & 55.50\% & 3.41 \\
& $k=4$ & 2.85$\times$ & 53.75\% & 5.30 \\
& $k=8$ & 4.15$\times$ & 60.62\% & 6.84 \\
& $k=16$ & 3.74$\times$ & 82.88\% & 4.51 \\
\midrule
 \multirow{4}{*}{Draft-SD} & $k=2$ & 0.99$\times$ & 28.22\% & 3.51 \\
& $k=4$ & 0.86$\times$ & 41.46\% & 2.07 \\
& $k=8$ & 0.66$\times$ & 59.84\% & 1.10 \\
& $k=16$ & 0.43$\times$ & 75.66\% & 0.57 \\
\midrule
 APEX-S & adaptive & 5.24$\times$ & 47.00\% & 11.15 \\
APEX-B & adaptive & 5.16$\times$ & 44.00\% & 11.72 \\
APEX-E & adaptive & 2.43$\times$ & 72.90\% & 3.33 \\
\bottomrule
\end{tabular}
\end{table*}

\begin{table*}[t]
\centering
\caption{Workload-level speed--waste comparison for \textbf{long-context completion}.  }
\label{tab:workload-long-context-completion}
\begin{tabular}{lcccc}
\toprule
 \textbf{Method} & \textbf{Depth} & \textbf{Speedup vs. AR $\uparrow$}  & \textbf{Wasted Token \% $\downarrow$} & \textbf{WAS $\uparrow$}\\
\midrule
\multirow{4}{*}{EAGLE-3} & $k=2$ & 1.59$\times$ & 70.50\% & 2.26 \\
 & $k=4$ & 1.89$\times$ & 77.75\% & 2.43 \\
& $k=8$ & 0.51$\times$ & 99.02\% & 0.52 \\
 & $k=16$ & 1.26$\times$ & 98.38\% & 1.28 \\
 \midrule
\multirow{4}{*}{n-gram SD} & $k=2$ & 1.22$\times$ & 37.41\% & 3.26 \\
 & $k=4$ & 3.32$\times$ & 34.50\% & 9.62 \\
 & $k=8$ & 2.03$\times$ & 61.53\% & 3.30 \\
 & $k=16$ & 6.40$\times$ & 66.25\% & 9.66 \\
 \midrule
\multirow{4}{*}{Draft-SD} & $k=2$ & 0.89$\times$ & 38.16\% & 2.34 \\
& $k=4$ & 1.08$\times$ & 33.24\% & 3.24 \\
 & $k=8$ & 1.07$\times$ & 25.68\% & 4.17 \\
& $k=16$ & 0.87$\times$ & 41.10\% & 2.12 \\
\midrule
APEX-S & adaptive & 5.04$\times$ & 49.50\% & 10.18 \\
APEX-B & adaptive & 1.83$\times$ & 44.42\% & 4.12 \\
APEX-E & adaptive & 2.72$\times$ & 41.05\% & 6.63 \\
\bottomrule
\end{tabular}
\end{table*}

\begin{table*}[t]
\centering
\caption{Workload-level speed--waste comparison for \textbf{long-chain reasoning}.  }
\label{tab:workload-long-chain-reasoning}
\begin{tabular}{lcccc}
\toprule
\textbf{Method} & \textbf{Depth} & \textbf{Speedup vs. AR $\uparrow$}  & \textbf{Wasted Token \% $\downarrow$} & \textbf{WAS $\uparrow$}\\
\midrule
\multirow{4}{*}{EAGLE-3} & $k=2$ & 1.46$\times$ & 77.00\% & 1.90 \\
 & $k=4$ & 2.13$\times$ & 71.75\% & 2.97 \\
& $k=8$ & 1.64$\times$ & 92.00\% & 1.78 \\
 & $k=16$ & 1.37$\times$ & 97.69\% & 1.40 \\
 \midrule
\multirow{4}{*}{n-gram SD} & $k=2$ & 1.70$\times$ & 65.00\% & 2.62 \\
 & $k=4$ & 2.46$\times$ & 63.50\% & 3.87 \\
 & $k=8$ & 3.57$\times$ & 67.88\% & 5.26 \\
 & $k=16$ & 4.57$\times$ & 77.69\% & 5.88 \\
 \midrule
\multirow{4}{*}{Draft-SD} & $k=2$ & 0.98$\times$ & 27.79\% & 3.53 \\
 & $k=4$ & 0.84$\times$ & 41.58\% & 2.02 \\
 & $k=8$ & 0.68$\times$ & 59.04\% & 1.15 \\
& $k=16$ & 0.42$\times$ & 75.18\% & 0.56 \\
\midrule
APEX-S & adaptive & 3.92$\times$ & 63.50\% & 6.17 \\
APEX-B & adaptive & 3.07$\times$ & 40.59\% & 7.56 \\
APEX-E & adaptive & 1.89$\times$ & 58.12\% & 3.25 \\
\bottomrule
\end{tabular}
\end{table*}

\begin{table*}[t]
\centering
\caption{Workload-level speed--waste comparison for \textbf{conversational generation}.  }
\label{tab:workload-conversational-generation}
\begin{tabular}{lcccc}
\toprule
 \textbf{Method} & \textbf{Depth} & \textbf{Speedup vs. AR $\uparrow$}  & \textbf{Wasted Token \% $\downarrow$} & \textbf{WAS $\uparrow$}\\
\midrule
\multirow{4}{*}{EAGLE-3} & $k=2$ & 1.78$\times$ & 61.00\% & 2.92 \\
 & $k=4$ & 2.35$\times$ & 66.25\% & 3.55 \\
 & $k=8$ & 2.61$\times$ & 79.88\% & 3.27 \\
& $k=16$ & 1.21$\times$ & 98.69\% & 1.23 \\
\midrule
\multirow{4}{*}{n-gram SD} & $k=2$ & 1.84$\times$ & 58.00\% & 3.17 \\
 & $k=4$ & 3.05$\times$ & 48.75\% & 6.26 \\
 & $k=8$ & 4.63$\times$ & 54.63\% & 8.48 \\
 & $k=16$ & 3.20$\times$ & 86.25\% & 3.71 \\
 \midrule
\multirow{4}{*}{Draft-SD} & $k=2$ & 0.91$\times$ & 29.03\% & 3.13 \\
& $k=4$ & 0.75$\times$ & 43.01\% & 1.74 \\
 & $k=8$ & 0.58$\times$ & 61.95\% & 0.94 \\
 & $k=16$ & 0.36$\times$ & 77.78\% & 0.46 \\
 \midrule
APEX-S & adaptive & 4.88$\times$ & 51.50\% & 9.48 \\
APEX-B & adaptive & 4.15$\times$ & 46.56\% & 8.91 \\
APEX-E & adaptive & 1.29$\times$ & 74.03\% & 1.74 \\
\bottomrule
\end{tabular}
\end{table*}

\begin{table*}[t]
\centering
\caption{Workload-level speed--waste comparison for \textbf{swe-style} tasks.  }
\label{tab:workload-swe-style-tasks}
\begin{tabular}{lcccc}
\toprule
\textbf{Method} & \textbf{Depth} & \textbf{Speedup vs. AR $\uparrow$}  & \textbf{Wasted Token \% $\downarrow$} & \textbf{WAS $\uparrow$}\\
\midrule
\multirow{4}{*}{EAGLE-3} & $k=2$ & 1.67$\times$ & 66.50\% & 2.51 \\
 & $k=4$ & 2.45$\times$ & 63.75\% & 3.84 \\
 & $k=8$ & 0.98$\times$ & 99.39\% & 0.99 \\
& $k=16$ & 1.24$\times$ & 98.50\% & 1.26 \\
\midrule
\multirow{4}{*}{n-gram SD} & $k=2$ & 1.76$\times$ & 62.00\% & 2.84 \\
 & $k=4$ & 2.50$\times$ & 62.50\% & 4.00 \\
& $k=8$ & 2.62$\times$ & 76.53\% & 3.42 \\
 & $k=16$ & 4.15$\times$ & 80.31\% & 5.17 \\
 \midrule
\multirow{4}{*}{Draft-SD} & $k=2$ & 1.02$\times$ & 20.47\% & 4.98 \\
& $k=4$ & 0.90$\times$ & 32.70\% & 2.75 \\
 & $k=8$ & 0.81$\times$ & 46.89\% & 1.73 \\
& $k=16$ & 0.54$\times$ & 66.06\% & 0.82 \\
\midrule
APEX-S & adaptive & 3.15$\times$ & 48.12\% & 6.55 \\
APEX-B & adaptive & 2.54$\times$ & 51.21\% & 4.96 \\
APEX-E & adaptive & 2.17$\times$ & 50.23\% & 4.32 \\
\bottomrule
\end{tabular}
\end{table*}

\subsection{Performance Across Workloads}
\label{sec:workload-results}
\label{ksweep}

The workload-level results reveal where adaptive control improves both speed and utilization, where it offers a trade-off, and where a fixed configuration remains preferable.

\paragraph{Mathematical reasoning.}
Table~\ref{tab:workload-mathematical-reasoning} shows the strongest acceleration achieved by \apex. APEX-S reaches $5.24\times$ speedup with $47.00\%$ waste, while APEX-B achieves $5.16\times$ speedup with $44.00\%$ waste. Both improve speed and reduce waste relative to the fastest fixed baseline, n-gram speculation at $k=8$, which achieves $4.15\times$ speedup with $60.62\%$ waste. APEX-B obtains the highest WAS of $11.72$, retaining nearly the acceleration of APEX-S while discarding fewer draft tokens.

\paragraph{Conversational generation.}
A similar improvement appears in Table~\ref{tab:workload-conversational-generation}. APEX-S achieves $4.88\times$ speedup with $51.50\%$ waste, compared with $4.63\times$ and $54.63\%$ for n-gram speculation at $k=8$. APEX-B reduces waste further to $46.56\%$ while achieving $4.15\times$ speedup. Increasing the fixed n-gram depth to $k=16$ worsens both speed and waste, showing that deeper speculation is not uniformly beneficial even within a workload where n-gram proposals are effective.

\paragraph{Long-context completion.}
Table~\ref{tab:workload-long-context-completion} presents a different trade-off. N-gram speculation at $k=16$ remains the fastest configuration at $6.40\times$ speedup with $66.25\%$ waste. APEX-S achieves $5.04\times$ speedup while reducing waste to $49.50\%$, yielding a higher WAS of $10.18$ versus $9.66$. This benefit does not extend to every learned profile: n-gram speculation at $k=4$ is both faster and less wasteful than APEX-B and APEX-E.

\paragraph{Long-chain reasoning.}
In Table~\ref{tab:workload-long-chain-reasoning}, n-gram speculation at $k=16$ achieves the highest speedup of $4.57\times$ but wastes $77.69\%$ of drafted tokens. APEX-B achieves $3.07\times$ speedup with $40.59\%$ waste, obtaining the highest WAS of $7.56$. Here, adaptive control offers substantially better draft-token utilization while retaining acceleration over autoregressive decoding.

\paragraph{Software-engineering tasks.}
Table~\ref{tab:workload-swe-style-tasks} shows that APEX-S achieves $3.15\times$ speedup with $48.12\%$ waste. N-gram speculation at $k=16$ is faster at $4.15\times$, but its waste rises to $80.31\%$. APEX-S consequently achieves the highest WAS of $6.55$, compared with $5.17$ for that baseline. It also achieves both higher speedup and lower waste than APEX-B and APEX-E on this workload.

\paragraph{Code generation.}
Code generation exposes a limitation of the learned controller (Table~\ref{tab:workload-code-generation}). N-gram speculation at $k=16$ achieves $4.25\times$ speedup and a WAS of $5.33$, exceeding all three \apex profiles on these metrics. The limitation also appears in direct speed--waste comparisons: n-gram speculation at $k=8$ is both faster and less wasteful than APEX-S, while EAGLE-3 at $k=4$ is both faster and less wasteful than APEX-B. These results identify code generation as a setting where improved routing and utility calibration are needed.

\begin{table}[t]
\centering
\caption{
\textbf{Ablation study across operating profiles}. Each variant removes one component
from the corresponding full \apex controller. Speedup is measured relative to
autoregressive decoding AR=1.00$\times$
}
\label{tab:ablations}
{
\begin{tabular}{llccc}
\toprule
\textbf{Profile} &
\textbf{Variant} &
\textbf{Speedup}  &
\textbf{Waste $\%$} &
\textbf{WAS} \\
\midrule

\multirow{6}{*}{\textbf{APEX-S}}
& \textbf{Full APEX-S}
& 2.44 $\times$
& 59.22 $\%$
& 4.12 \\

& w/o survival loss
& 1.62 $\times$
& 68.16 $\%$
& 2.37 \\

& w/o cost/TPS objective
& 2.41 $\times$
& 62.69\%
& 3.85 \\

& w/o entropy features
& 2.49 $\times$
& 62.41 $\%$ 
& 3.99 \\

& w/o repetition features
& 2.42 $\times$
& 62.22 $\%$
& 3.89 \\

& w/o verifier-history features
& 1.14 $\times$
& 54.74 $\%$
& 2.09\\

\midrule

\multirow{6}{*}{\textbf{APEX-E}}
& \textbf{Full APEX-E}
& 2.34 $\times$
& 52.85 $\%$
& 4.43 \\

& w/o survival loss
& 1.56 $\times$
& 67.62 $\%$
& 2.31 \\

& w/o cost/TPS objective
& 2.26 $\times$
& 57.83 $\%$
& 3.91 \\

& w/o entropy features
& 2.31 $\times$
& 58.06 $\%$
& 3.98 \\

& w/o repetition features
& 2.29 $\times$
& 58.12 $\%$
&  3.94\\

& w/o verifier-history features
& 1.31 $\times$
& 56.48 $\%$
& 2.32\\

\midrule

\multirow{6}{*}{\textbf{APEX-B}}
& \textbf{Full APEX-B}
& 2.40 $\times$
& 54.62 $\%$
& 4.39 \\

& w/o survival loss
& 1.53 $\times$
& 69.38 $\%$
& 2.21 \\

& w/o cost/TPS objective
& 2.33 $\times$
& 57.83 $\%$
& 4.03 \\

& w/o entropy features
& 2.35 $\times$
& 58.34 $\%$
& 4.03 \\

& w/o repetition features
& 2.30 $\times$
& 58.45 $\%$
& 3.94 \\

& w/o verifier-history features
& 1.14 $\times$
& 56.22 $\%$
& 2.03  \\

\bottomrule
\end{tabular}
}
\end{table}

\begin{table*}[t]
\centering
\caption{\apex candidate-\(k\) sweep under the quality-balanced operating profile. Higher speedup and WAS are better; lower wasted-token percentage is better.}
\label{tab:apex-ksweep-quality-balanced}
\begin{tabular}{llcccc}
\toprule
\textbf{Method} & \textbf{Candidate \(k\) set} & \textbf{Mean \(k\)} & \textbf{Speedup vs. AR} & \textbf{Wasted Token \%} & \textbf{WAS} \\
\midrule
\multirow{3}{*}{\textbf{APEX-S}} & \(k=1\text{--}16\) & 3.61 & 3.34$\times$ & 47.18\% & 7.07 \\
 & \(k=3\text{--}16\) & 4.54 & 3.43$\times$ & 50.57\% & 6.79 \\
 & \(k\in\{2,4,8,12,16\}\) & 3.92 & 3.39$\times$ & 47.95\% & 7.06 \\
\midrule
\multirow{3}{*}{\textbf{APEX-B}}& \(k=1\text{--}16\) & 3.09 & 3.25$\times$ & 42.76\% & 7.60 \\
 & \(k=3\text{--}16\) & 4.28 & 3.34$\times$ & 48.55\% & 6.89 \\
 & \(k\in\{2,4,8,12,16\}\) & 3.57 & 3.31$\times$ & 45.10\% & 7.33 \\
\midrule
\multirow{3}{*}{\textbf{APEX-E}} & \(k=1\text{--}16\) & 2.96 & 1.56$\times$ & 59.88\% & 2.61 \\
& \(k=3\text{--}16\) & 4.13 & 1.60$\times$ & 67.83\% & 2.35 \\
& \(k\in\{2,4,8,12,16\}\) & 3.40 & 1.58$\times$ & 62.85\% & 2.52 \\
\bottomrule
\end{tabular}
\end{table*}

\subsection{Ablation Study}
\label{sec:ablation}

We examine the contributions of survival supervision, cost and throughput supervision, and decoding-state features in Table~\ref{tab:ablations}. This study uses a smaller diagnostic subset containing twenty requests from each of the six workloads, for a total of $120$ requests. All variants are evaluated on the same requests under the same serving configuration. Ablation effects are measured relative to the corresponding full controller on this subset; the absolute values therefore differ from the main evaluation.

\paragraph{Full-controller profiles.}
Among the full controllers, APEX-S achieves the highest speedup of $2.44\times$ with $59.22\%$ waste. APEX-B achieves $2.40\times$ speedup with $54.62\%$ waste, while APEX-E achieves $2.34\times$ speedup with $52.85\%$ waste and the highest WAS of $4.43$. The profile ordering differs from the aggregate evaluation, further illustrating its dependence on the evaluated requests.

\paragraph{Survival supervision.}
Removing the survival loss produces the largest increase in waste among the examined ablations for all three profiles. Waste rises to $67.62\%$--$69.38\%$, speedup falls to $1.53\times$--$1.62\times$, and WAS decreases to $2.21$--$2.37$. These results support the role of first-rejection survival supervision in selecting depths that convert drafted tokens into accepted progress.

\paragraph{Cost and throughput supervision.}
Removing the cost/TPS objective causes smaller speed reductions but consistently increases waste and lowers WAS. For APEX-B, waste increases from $54.62\%$ to $57.83\%$, while WAS decreases from $4.39$ to $4.03$. The results support accounting for runtime cost alongside acceptance behavior when comparing candidate actions.

\paragraph{Entropy and repetition features.}
Removing either feature group increases waste and reduces WAS across all three profiles. The effect on raw speedup is smaller and is not uniformly negative. For example, removing entropy features from APEX-S increases speedup from $2.44\times$ to $2.49\times$, but also increases waste from $59.22\%$ to $62.41\%$, lowering WAS from $4.12$ to $3.99$. These features therefore contribute primarily to the observed speed--waste balance.

\paragraph{Verifier-history features.}
Removing recent verifier feedback produces the largest speed reductions, with speedup falling to $1.14\times$--$1.31\times$ and WAS to $2.03$--$2.32$. For APEX-S, waste actually decreases from $59.22\%$ to $54.74\%$, but speedup falls from $2.44\times$ to $1.14\times$. This illustrates why lower waste alone is insufficient and supports using recent acceptance and rejection outcomes to guide block-level decisions.

\subsection{Sensitivity to Candidate Depths}
\label{sec:depth-sensitivity}

Table~\ref{tab:apex-ksweep-quality-balanced} reports a separate comparison of three candidate-depth sets across all \apex profiles:
$\mathcal{K}_{\mathrm{dense}}=\{1,\ldots,16\}$,
$\mathcal{K}_{\mathrm{deep}}=\{3,\ldots,16\}$, and
$\mathcal{K}_{\mathrm{sparse}}=\{2,4,8,12,16\}$.
The dense set permits fine-grained choices, the deep set removes the two shallowest actions, and the sparse set reduces the number of available depths.

Removing shallow actions increases the mean selected depth and slightly improves speedup across all three profiles, but also increases waste and reduces WAS. For APEX-B, moving from the dense to the deep set raises speedup from $3.25\times$ to $3.34\times$, while waste increases from $42.76\%$ to $48.55\%$ and WAS decreases from $7.60$ to $6.89$. Shallow actions therefore contribute to controlling waste even when excluding them produces a modest speed gain.

The sparse set provides an intermediate operating point. For APEX-S, it achieves $3.39\times$ speedup with $47.95\%$ waste, compared with $3.34\times$ and $47.18\%$ for the dense set. Their WAS values are nearly identical at $7.06$ and $7.07$. APEX-B and APEX-E also retain acceleration under the sparse set, although their WAS values remain below those obtained with the dense set. These results show that fine-grained depth choices are not equally important for every profile, while access to shallow actions consistently improves the measured speed--waste balance in this study.

\subsection{Discussion}

The results support adaptive speculative control while establishing its workload-dependent limits. APEX-S provides the highest aggregate speedup, and APEX-B offers lower aggregate waste at reduced acceleration. Mathematical reasoning and conversational generation show simultaneous improvements over the fastest fixed baselines, while other workloads expose trade-offs or favor fixed configurations. The ablations support combining survival supervision, runtime-cost supervision, and recent verifier feedback. Together, these findings motivate evaluating speculative controllers through both latency and draft-token utilization, with profile selection guided by the workload and the desired operating point.

\section{Conclusion and Future Work}
\label{sec:conclusion}

\apex combines request-level speculator selection with block-level depth adaptation to balance decoding speed and draft-token waste. Its learned controller provides a modular way to coordinate heterogeneous proposal mechanisms while preserving target-model verification. The results demonstrate substantial acceleration with lower waste than aggressive fixed-depth configurations, with trade-offs that vary across workloads.

Future work will use traces collected during decoding to update the controller online as workloads and execution conditions change. We also plan to evaluate \apex beyond single-turn prompts, including multi-turn conversations, reinforcement-learning rollouts, and tool-augmented generation. These settings would test how well the controller responds to changes in context and generation patterns over successive steps. To expand beyond the current three speculative methods, we plan to add a common interface for new speculators and their execution traces. The router and depth controller could then be retrained on these traces to incorporate the additional methods.

\bibliography{iclr2027_conference}
\bibliographystyle{iclr2027_conference}

\end{document}